\documentclass[preprint,review,12pt]{elsarticle}
\usepackage{amssymb}
\usepackage{amsthm}
\usepackage[figuresright]{rotating}
\usepackage{amsmath}
\usepackage{mathrsfs}
\usepackage{newtxmath}
\usepackage{color, colortbl}
\usepackage{amsfonts}
\usepackage{mathtools}
\usepackage{algorithm}
\usepackage{algorithmic}
\usepackage{soul}
\usepackage{url}
\usepackage[utf8]{inputenc}
\usepackage{caption}
\usepackage{subcaption}
\usepackage{graphicx}
\usepackage{booktabs}
\usepackage{xcolor}
\usepackage{multirow}
\usepackage{verbatim}
\usepackage{float}
\usepackage{textcomp}
\usepackage{lipsum}
\usepackage{stfloats}
\usepackage{array}
\usepackage{comment}
\usepackage{makecell}

\usepackage{setspace} 

\usepackage{hyperref}
\hypersetup{
    colorlinks=true,
    linkcolor=blue,
}
\journal{Pattern Recognition}
\begin{document}

\begin{frontmatter}

\title{\textit{Percept}, \textit{Chat}, \textit{Adapt}: \\
Knowledge Transfer of Foundation Models for Open-World Video Recognition}

\author[inst1,inst2]{Boyu Chen}
\ead{chenboyu18@mails.ucas.ac.cn}
\author[inst1,inst2]{Siran Chen}
\ead{sr.chen@siat.ac.cn}
\author[inst1,inst2,inst3]{Kunchang Li}
\ead{kc.li@siat.ac.cn}
\author[inst1,inst2]{Qinglin Xu}
\ead{ql.xu@siat.ac.cn}
\author[inst3]{Yu Qiao}
\ead{yu.qiao@siat.ac.cn}
\author[inst1,inst3]{Yali Wang\corref{cor1}}
\cortext[cor1]{Corresponding author}
\ead{yl.wang@siat.ac.cn}

\affiliation[inst1]{organization={Shenzhen Institute of Advanced Technology, Chinese Academy of Sciences},
            city={Shenzhen},
            country={China}}
\affiliation[inst2]{organization={the School of Artificial Intelligence, University of Chinese Academy of Sciences},
            city={Beijing},
            country={China}}
\affiliation[inst3]{organization={Shanghai AI Laboratory},
            city={Shanghai},
            country={China}}

\begin{abstract}
Open-world video recognition is challenging since traditional networks are not generalized well on complex environment variations. 
Alternatively, foundation models with rich knowledge have recently shown their generalization power. 
However, how to apply such knowledge has not been fully explored for open-world video recognition. 
To this end, we propose a generic knowledge transfer pipeline, which progressively exploits and integrates external multimodal knowledge from foundation models to boost open-world video recognition.
We name it \textbf{PCA}, based on three stages of \textbf{P}ercept, \textbf{C}hat, and \textbf{A}dapt. 
First, we perform Percept process to reduce the video domain gap and obtain external visual knowledge. Second, we generate rich linguistic semantics as external textual knowledge in Chat stage. 
Finally, we blend external multimodal knowledge in Adapt stage, by inserting multimodal knowledge adaptation modules into networks. 
We conduct extensive experiments on three challenging open-world video benchmarks, i.e., TinyVIRAT, ARID, and QV-Pipe. Our approach achieves state-of-the-art performance on all three datasets. 
\end{abstract}


\end{frontmatter}

\section{Introduction}

Video recognition tasks are important in multimedia research,
due to their wide applications in surveillance, robotics, and so on.
With the rapid development of deep learning,
a number of video networks have been proposed in recent years\cite{uniformerv2, pr-arclip, chen2022low, chen2025g, chen2025lvagent, chen2025super}.
However,
these recognition models mainly work on the classical video benchmarks collected under desirable conditions.
In many realistic applications,
the camera-shooting environments are much more complex with 
low-resolution targets, 
bad illumination conditions, 
unusual video scenes, 
etc.
In such open-world scenarios,
most existing video models are not well generalized due to the lack of external domain knowledge.

\begin{figure*}[t]
   \centering
   \includegraphics[width=\textwidth]{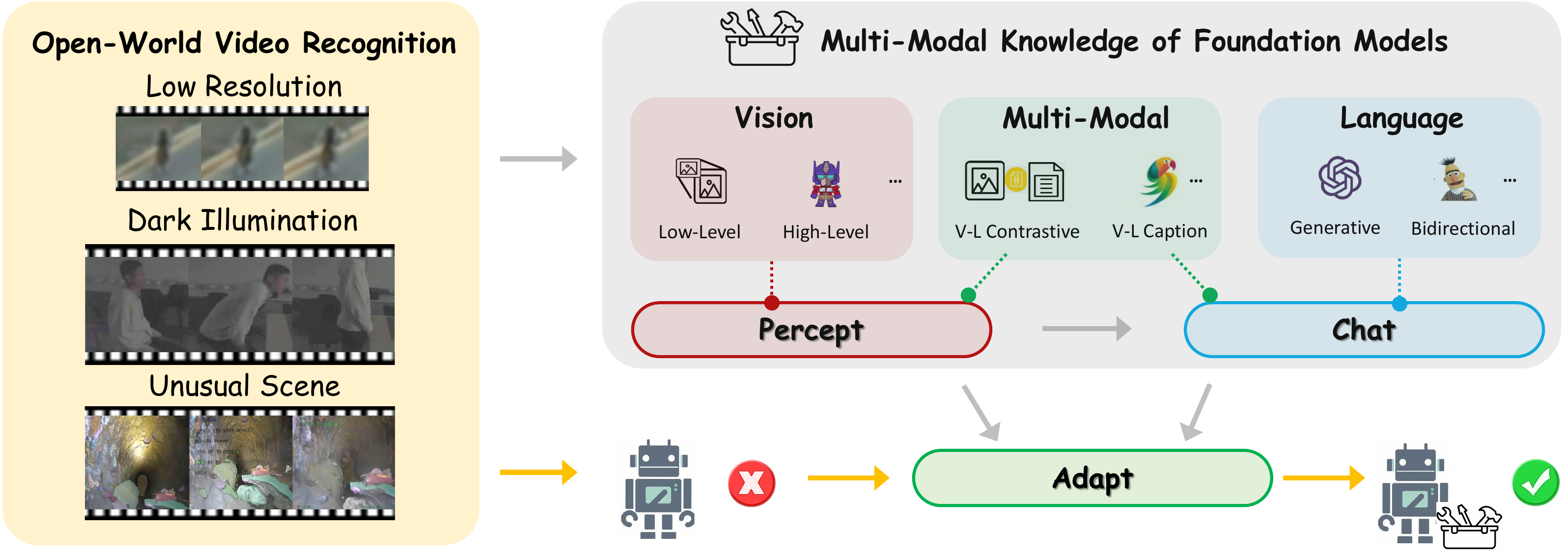}
   \caption{\textbf{Overview of PCA. }Given an open-world video, we first perform percept to reduce the domain gap of the video and obtain enhanced visual features. Then, we use large language models and large visual-language models for chat process to obtain external textual knowledge. Finally, in the adapt process, we integrate multimodal knowledge into the training process, enhancing the models' ability for open-world video perception.}
\label{fig:pcaIntro}
\end{figure*}

Alternatively,
the remarkable success of large language models\cite{llama, chatgpt, vicuna, Instructgpt} have driven the explosive development of various foundation models\cite{videochat, blip,  internvideo, llava}.
Based on web-scale pre-training,
these foundation models contain diversified semantic knowledge
and can be adapted for low-shot generalization\cite{flamingo,chen2025top,chen2025videochat,wang2025videochat,chen2025vragent, yue2025uniflow}.
However,
how to leverage such knowledge of these models for open-world video recognition has not been fully explored.
For this reason,
we propose a generic knowledge transfer paradigm,
namely \textit{PCA},
which can progressively exploit and integrate external knowledge of open-world videos from foundation models,
by three cascade stages of \textit{Percept}, \textit{Chat}, \textit{Adapt}.
As shown in Fig.\ref{fig:pcaIntro},
we treat the foundation models as a knowledge container of open-world videos,
where we build \textit{PCA} with the prior knowledge learned from vision, language, and multimodal models.

In the \textit{Percept} stage,
we first enhance poor-conditioned videos for domain gap reduction,
through low-level visual models (e.g., RealBasicVSR\cite{realbasicvsr}).
Then,
we extract external visual knowledge of enhanced videos,
from high-level visual models (e.g., UniFormer\cite{uniformer}) or visual encoders of backbones (e.g. CLIP\cite{clip}).
In the \textit{Chat} stage,
we further generate textual descriptions about predicted labels or videos,
by chatting with large language models (e.g., ChatGPT\cite{chatgpt}) or visual captioners of multimodal language models (e.g., VideoChat\cite{videochat}).
Such diversified descriptions are used as external textual knowledge about videos.
Finally,
we introduce a distinct multimodal knowledge adaptation module in the \textit{Adapt} stage,
which flexibly blends various complementary knowledge into vision backbones for boosting open-world video recognition.

The contributions of our work are summarized as following three folds:

\begin{itemize}
\item We propose a generic knowledge transfer paradigm \textit{PCA},
which progressively mines various external knowledge from foundation models,
and boosts open-world video recognition by \textit{Percept}, \textit{Chat}, \textit{Adapt}.
\item We introduce a plug-and-play knowledge adaptation module,
allowing us to integrate \textit{PCA} flexibly into various video backbones for open-world recognition.
\item We evaluate \textit{PCA} on three challenging tasks of open-world video recognition,
e.g.,
low-resolution action recognition on TinyVIRAT\cite{tinyvirat}, 
dark-light action recognition on ARID\cite{arid}, 
and pipe anomaly detection on QV-Pipe\cite{videopipe}.
Extensive experiments show that
our method simply achieves state-of-the-art performance on all these open-world video datasets.
\end{itemize}

\section{Related work}
\noindent\textbf{Open World Video Recognition}.
In recent years, video recognition models\cite{vivit, timesformer, actionclip, uniformer} have entered a period of rapid development. 
However, these networks are primarily designed for academic scenarios such as Kinetics\cite{kinetics400} and Something-something\cite{ssv1} datasets, rather than real-world model structures.In real-world scenarios, the captured videos contain complex environment variations\cite{videopipe, tmmopenset, openworld_ar, tpami_tem_action_localization}. 
For example, in dark environments, the low light intensity obscures important details in the video, making the action recognition task challenging.  
The DarkLight network\cite{darklight} improves the accuracy of behavior recognition models in low-light conditions by incorporating a cross-attention interaction between the gamma-corrected features and the original image features. 
In real surveillance scenarios, when the subject is far from the camera, inevitable blurriness occurs. A weakly supervised attention mechanism is applied to highlight the regions with action\cite{tinyvirat}. 
QV-Pipe\cite{videopipe} introduces video perception tasks to industrial applications 
and 
addresses the problem of real-world city pipeline defect recognition.
In these open-world scenarios, the most common approach is the model ensemble\cite{videopipe} method with weighted addition. 
However, these methods can not generalize well in open-world videos with significant domain gaps. 

\noindent\textbf{Foundation Models}.
Recently, language foundation models have experienced rapid development and have achieved significant improvements in text generalization capabilities.
GPT-3\cite{gpt3} is a milestone model in language modeling that demonstrates the qualitative improvement brought about by increasing data and model parameters. 
Models such as InstructGPT\cite{Instructgpt}, LLAMA\cite{llama}, ChatGPT\cite{chatgpt}, and Vicuna\cite{vicuna} make it possible for models to engage in natural conversations with humans through scaling up parameters and instruction fine-tuning. 
Afterwards, the innovation of language model technology has also promoted the development of visual models.
The development of image foundation models also progressed rapidly. 
Using web-scale data and contrastive learning, image foundation models\cite{clip, beit3} have achieved excellent performance in various tasks. 
Video foundation models such as UniFormerV2\cite{uniformerv2}, and InternVideo\cite{internvideo} possess exceptional video understanding capabilities. 
Finally, visual and textual information are combined to generate multimodal foundation models\cite{clip,  flamingo}.
Models such as BLIP\cite{blip}, VideoChat\cite{videochat} and LLaVA\cite{llava} have already acquired the ability to jointly process visual and natural textual information.
However, how to apply the knowledge of these foundation models in open-world video understanding is still under exploration.

\noindent\textbf{Knowledge Adapters}. 
Adapter modules in deep learning networks play a crucial role in various fields like knowledge transfer, large-scale pre-training, and multimodal training. 
By incorporating an adapter module, deep learning models have abilities to transfer pre-trained knowledge to downstream tasks and multimodal tasks.
Models with adapters can be primarily categorized into two types: embedded adapter and external adapter. Embedded adapter involves modifying or adding modules within the transformer encoder\cite{vit}. UniFormer\cite{uniformer, uniformerv2}, ST-Adapter\cite{stadapter} and BEIT-3\cite{beit3}, on the other hand, have made modifications to the self-attention and MLP modules, improving the performance of the models on multiple tasks.

As for the external adapter, the original pre-trained large-scale model is preserved, and connectors are added externally for downstream tasks or other domains. CoOp\cite{CoOp} incorporated additional learnable parameters as adapters to the visual branch. 
Clip-Adapter\cite{clipadapter}, EVL\cite{frozenclip}, and Flamingo\cite{flamingo} freeze the backbone network and introduce adapter modules for efficient fine-tuning. However, the models with adapter modules are not well explored in real-world video perception scenarios that involve significant domain gaps.

\section{Methodology}
In this section,
we mainly introduce the proposed generic knowledge transfer pipeline PCA.
First,
we describe the general framework of PCA.
Then,
we will provide a detailed explanation of the three components of the PCA pipeline, 
namely Percept, Chat, and Adapt.
\subsection{Overview of PCA}

As shown in Fig.\ref{fig:pcaIntro}, we propose a general knowledge transfer pipeline, which serves as a knowledge container by integrating multimodal external knowledge into the training process. 
\textbf{PCA} pipeline consists of three cascaded stages: \textbf{P}ercept, \textbf{C}hat, and \textbf{A}dapt. 
First, the perception process utilizes a low-level visual model to enhance the raw video. Then, visual features are extracted from the enhanced video as external visual knowledge. 
Second, the chat process generates rich language semantics about open-world video as external textual knowledge. 
Finally, we introduce plug-and-play adapter modules
that seamlessly integrates external multimodal knowledge into the model training process
to assist with open-world video recognition.


\subsection{Percept}

\begin{figure*}[t]
  \includegraphics[width=0.95\textwidth]{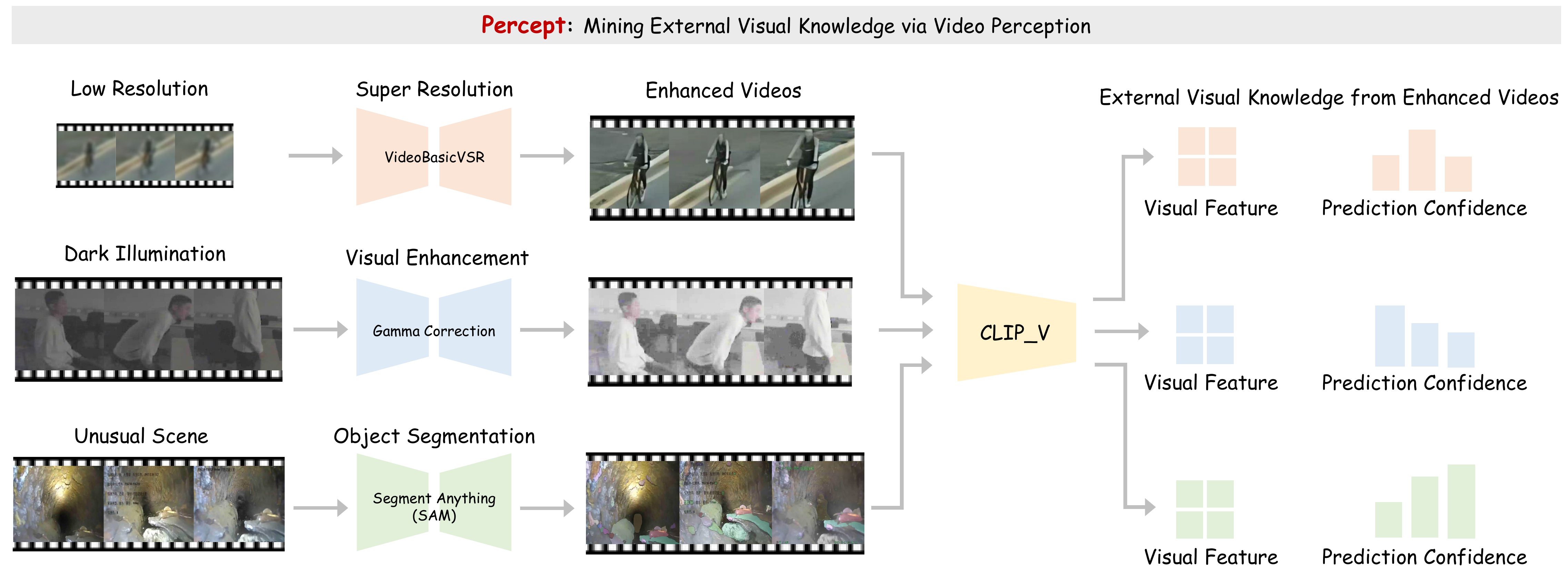}
  \caption{\textbf{Percept process.} We first preprocess open-world videos using low-level visual models to enhance the visual content and reduce the domain gap. Subsequently, we utilize visual networks to extract enhanced features and predict the confidence of all categories.}
  \label{fig:percept}
\end{figure*}

Due to the significant domain gaps in videos from realistic scenarios in the open world, traditional video models do not generalize well. To reduce the domain gap of open-world videos, we introduce a percept process to enhance the visual information in open-world videos. 
The process of the percept is illustrated in Fig.\ref{fig:percept}.
We first employ a visual enhancement model $\mathscr{F}$ to preprocess the given videos $V$, reducing the domain gap and enriching the visual information of open-world videos. The process can be formulated as 
\begin{equation}
\label{eq1}
\begin{aligned}
\widetilde{V}=\mathscr{F}(V);
\end{aligned}
\end{equation}
For instance, in the TinyVIRAT\cite{tinyvirat} real low-resolution scenes, in order to obtain clearer visual information, we employed the RealBasicVSR\cite{realbasicvsr} model to perform super-resolution on the videos and enhanced certain crucial visual details. 
To restore some indistinguishable details in the ARID nighttime dataset\cite{arid}, we brighten the video frames using gamma correction. 
The distinctive feature of the QV-Pipe\cite{videopipe} dataset lies in its significant domain gaps. 
To annotate the semantic regions of the videos and reduce the domain gap, we employed the Segment Anything\cite{segmentanything} model for preprocessing. 
By incorporating semantic segmentation masks into the original videos, we highlight some regions in the video data.
Based on the aforementioned method, we obtained visual enhanced video $\widetilde{V}$.
Then we utilize video foundation models $\mathscr{B}$ to extract enhanced open-world video features $F_V$ and predicted scores $S$. 
The predicted scores $S$ refer to the confidence of each category obtained after passing the visual foundation models. 
The whole process is formulated as
\begin{equation}
\label{eq2}
\begin{gathered}
\{F_V, S\}=\mathscr{B}(\widetilde{V}).
\end{gathered}
\end{equation}

It should be noted that the foundation models $\mathscr{B}$ can encompass various kinds of models, not just a single model. In order to validate the applicability of our PCA pipeline, we chose well-performing models in the field of video understanding as baselines $\mathscr{B}$. The selected models include the best-performing ones in these three real-world behavior recognition datasets, facilitating result comparison. 
The selection of these five models includes CNN, Transformer, and a combination of CNN and Transformer. We chose them to demonstrate that the PCA pipeline serves as a general knowledge transfer framework. 
The visual knowledge extractors include popular CNN models (R(2+1)D\cite{r2p1d}), 
transformer models (CLIP-ViT\cite{clip}, Timesformer\cite{timesformer}), 
as well as hybrid models (UniFormer\cite{uniformer}, CLIP-3D\cite{clip}) combining convolution and attention modules. 
Among them, CLIP-ViT is the visual branch of CLIP\cite{clip}. 
CLIP-3D\cite{clip} is obtained by inserting $3\times1\times1$ 3D-convolutions into the transformer blocks of CLIP-ViT for better temporal information aggregation, following UniFormerV2\cite{uniformerv2}.

\subsection{Chat}

\begin{figure*}[t]
  \includegraphics[width=0.95\textwidth]{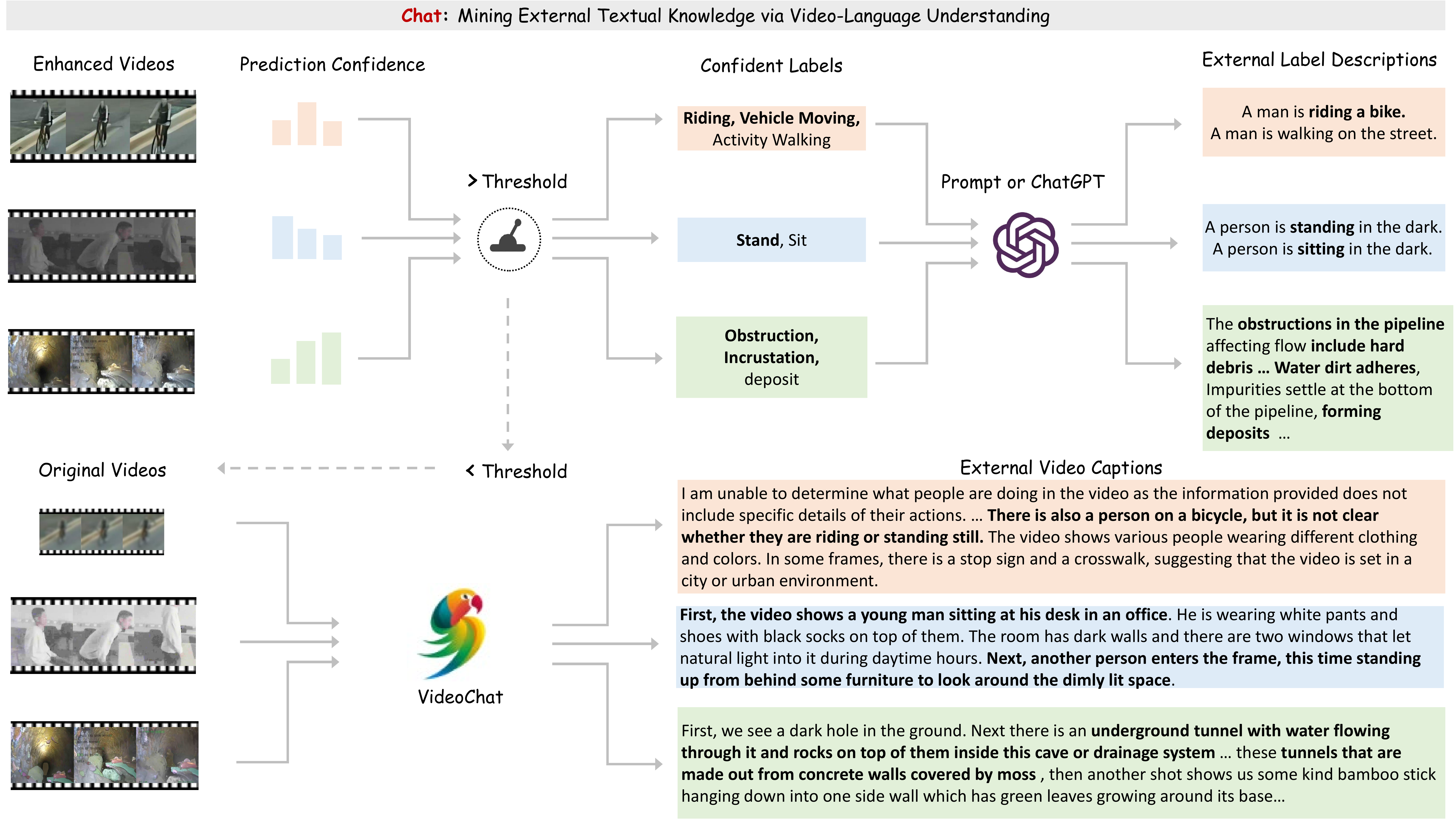}
  \centering
  \caption{\textbf{Chat process.} Due to the difficulty of recognizing open-world videos, text knowledge is needed to assist in video perception. 
  If the max prediction score from the percept process is higher than the set threshold, prompt methods are used to semantically expand the predicted labels. 
  If no category has a confidence score higher than the threshold, it indicates that the enhanced visual features are not applicable. In this case, VideoChat is used to obtain captions for the original video. Thus, external video captions and label descriptions are obtained.}
  \label{fig:chat}
\end{figure*}

Real-world scene videos with the complexity of environmental variables are difficult to perceive by pure visual features. 
For example, in QV-Pipe dataset, it is difficult to directly classify some categories like obstruction, incrustation, and deposit.
As shown in Fig.\ref{fig:chat}, through text prompts, the model can receive detailed label explanation knowledge like obstruction is caused by hard debris, water dirt adheres, etc.
With the help of external textual knowledge, the model can further indicate important clues based on the enhanced videos.
Therefore, we introduce the Chat process to extract external textual knowledge, which complements the visual features for open-world video perception. The entire chat process is formulated as
\begin{align}
\label{eq3}
T_p=\mathscr{P}(S) \qquad \max (S) \geqslant \sigma \\
T_c=\mathscr{C}(V) \qquad \max (S) < \sigma 
\end{align}

$T_p$ is the supplementary textual knowledge obtained from prompting. 
$T_c$ is the external video caption obtained by Videochat\cite{videochat}. 
We will explain how to generate them in the following.
Specifically,
to save computational resources required to invoke the visual language model and obtain effective external textual knowledge,
we introduced a threshold-controlled switch as shown in Fig.\ref{fig:chat}. This switch allows us to choose between prompt knowledge and caption knowledge from the large visual language model,
based on the confidence scores $S$ provided in the percept process. 
When the maximum confidence score exceeds the pre-defined threshold $\sigma$, 
it indicates that the augmented features yield meaningful results. 
In such cases, 
we use the prompt method $\mathscr{P}$ to obtain supplementary textual knowledge $T_p$. 
In the TinyVIRAT\cite{tinyvirat} and ARID\cite{arid} datasets, we semantically expand the predicted labels. 
We add a subject (e.g., "A man is") and adverbial phrases (e.g., "on the street", "in the dark") to the category labels. If there are multiple candidate labels, we concatenate the expanded texts together. 
The example is shown in Fig.\ref{fig:chat}. 
In QV-Pipe\cite{videopipe}, explanatory text is published for each category. 
For data with multiple labels, we use ChatGPT\cite{chatgpt} to integrate and summarize the explanations of these labels.

On the other hand, 
if none of the maximum confidence scores exceeds the threshold $\sigma$, 
it suggests that the percept model lacks confidence in the results obtained from visually enhanced videos. 
In such cases, 
we directly input the original video into the VideoChat\cite{videochat} $\mathscr{C}$, to obtain the external video caption $T_c$. 
Finally, we use BERT\cite{bert} model $\mathscr{L}$ to extract textual features $F_T$ from $T_p$ or $T_c$. $T_p$ is the supplementary textual knowledge obtained from the prompt method. The process can be formulated as
\begin{align}
\label{eq_bert}
F_T=\mathscr{L}(T_p) \quad or \quad F_T=\mathscr{L}(T_c)
\end{align}
By employing the approach mentioned above, we obtain features that incorporate external textual knowledge.

\subsection{Adapt}

\begin{figure*}[t]
  \includegraphics[width=0.95\textwidth]{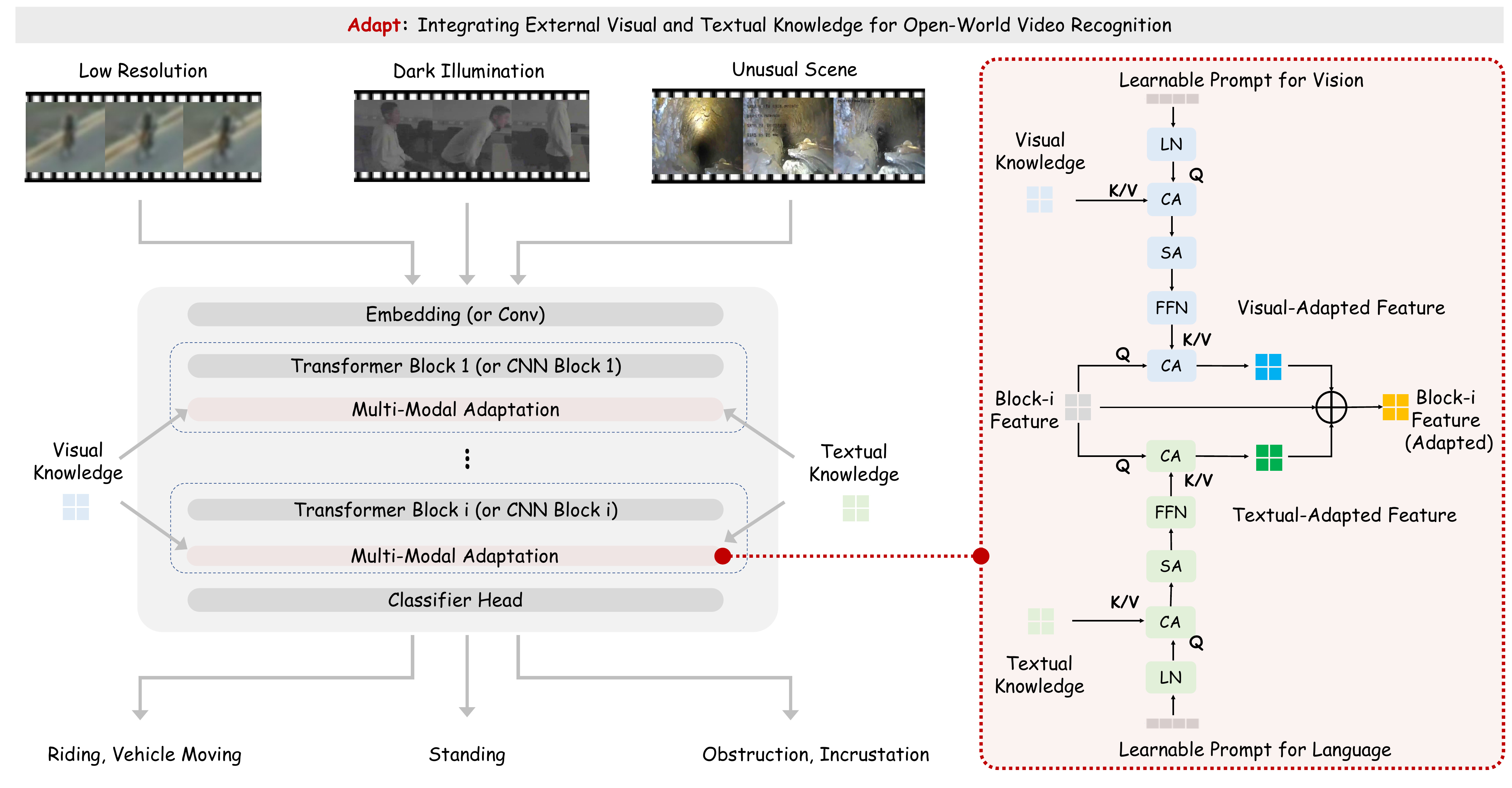}
  \caption{\textbf{Adapt process.} We incorporate visual and textual knowledge into the training process through specially designed Adapt modules. The Adapt module is capable of integrating multimodal knowledge and can be seamlessly inserted into any block of the networks, enabling plug-and-play usage.}
  \label{fig:adapt}
\end{figure*}

To incorporate external visual features $F_V$ and textual features $F_T$ into the open-world video training process, we introduced the adapt process as shown in Fig.\ref{fig:adapt}. Building upon the pre-trained foundation model, we embedded adapt modules into the intermediate layers. 
To align and adapt external knowledge with the network, we introduce learnable prompts and attention mechanisms to integrate external knowledge.
The compressed knowledge $\widetilde{F}_T$ and $\widetilde{F}_V$ is obtained by cross-attention $\mathcal{CA}$ with normed learnable prompts ${P}_L$.
$\mathcal{SA}$ and $\mathcal{FFN}$ are used to increase the complexity of the network and incorporate some non-linear knowledge. The process can be formulated as
\begin{equation}
\label{eq6}
\begin{aligned}
& \widetilde{F}_T=\mathcal{FFN}\left(\mathcal{SA}\left(\mathcal{CA}\left({P}_L, F_T\right)\right)\right) \\
& \widetilde{F}_V=\mathcal{FFN}\left(\mathcal{SA}\left(\mathcal{CA}\left({P}_L, F_V\right)\right)\right)
\end{aligned}
\end{equation}

Normed learnable parameters ${P}_L$ come from random initialization with a standard normal distribution following a layer norm process.
We utilize cross-attention to fuse the processed external knowledge with the intermediate layers of the network.
$F_B$ is the query, while $\widetilde{F}_T$ and $\widetilde{F}_V$ are the key-value inputs of the cross-attention. 
Through the above process, the textual-adapted features $G_T$ and the visual-adapted features $G_V$ are obtained. The process is formulated as
\begin{equation}
\label{eq5}
\begin{aligned}
& G_T=\mathcal{CA}\left(F_B, \widetilde{F}_T\right) \\
& G_V=\mathcal{CA}\left(F_B, \widetilde{F}_V\right) 
\end{aligned}
\end{equation} 
To fuse the external knowledge into the network, we integrate textual-adapted features $G_T$ 
and visual-adapted features $G_V$ with the intermediate layer features $F_B$ of the foundation model by weighted addition. This process is formulated as
\begin{equation}
\label{eq4}
\widetilde{F}_B=F_B+w_1 \cdot G_V +w_2 \cdot G_T
\end{equation}

\begin{figure*}[t]
  \includegraphics[width=\textwidth]{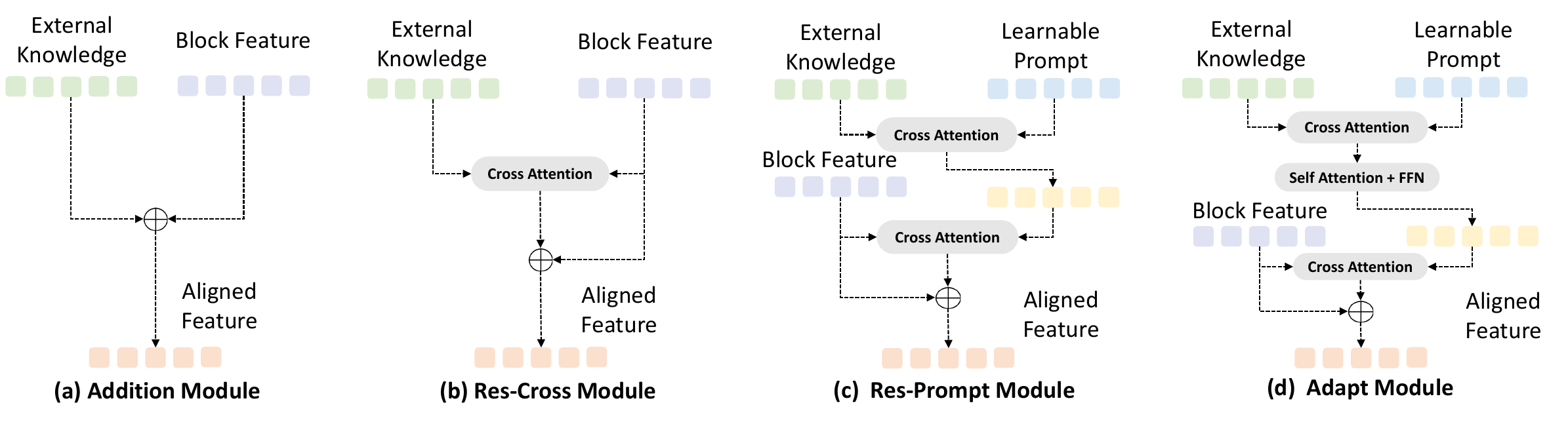}
  \centering
  \caption{\textbf{Structure Variants}. We explore four fusion structures to integrate external knowledge with the model. 
  Fig.(a) represents the direct weighted addition. 
  Fig.(b) builds upon cross-attention and incorporates a residual module. 
  Fig.(c) involves the integration of external knowledge and learnable prompts based on (b). Fig.(d) represents the adapt module which adds self-attention and FFN based on (c). }
  \label{fig:structures}
\end{figure*}

To further validate the superiority of the Adapt module over other classic fusion methods, we tested four different structures. 
Fig.\ref{fig:structures}(a) represents the direct weighted addition of external knowledge and block feature. 
Fig.\ref{fig:structures}(b) builds on cross-attention and incorporates a residual module. 
The above two structures are common methods for feature fusion.
In order to reduce the dimension of features and redundant external knowledge, we introduced learnable prompts to the adapter modules.
Fig.\ref{fig:structures}(c) involves integrating external knowledge and learnable prompts based on (b). 
To achieve a deeper level of fusion between external knowledge and learnable prompts, we utilize self-attention and FFN based on (c). 
Fig.\ref{fig:structures}(d) represents our adapt module.
The experiment shown in Table \ref{tab:structures} demonstrated that the Adapt module in Fig.\ref{fig:structures}(d) outperformed the other three structures.

\section{Experiments}

\noindent\textbf{Datasets.} We perform extensive experiments on the open-world video perception benchmarks including
TinyVIRAT\cite{tinyvirat},
ARID\cite{arid},
and QV-Pipe\cite{videopipe}.
TinyVIRAT dataset\cite{tinyvirat} is a real-world low-resolution action recognition dataset. 
The resolution of the videos in the dataset ranges from 10 to 128 pixels. There are a total of 26 categories, and each video may have multiple labels.
ARID dataset\cite{arid} is specifically introduced for dark light scenes action recognition, consisting of 3,780 video clips categorized into 11 action classes. 
QV-Pipe dataset\cite{videopipe} is a specialized dataset for anomaly detection in underground water pipelines in urban areas. It is a unique dataset that exhibits significant domain differences compared to ordinary classification datasets. There are a total of 17 categories, consisting of 1 normal class and 16 defect classes.
Note that,
our PCA is a generic knowledge transfer framework which aims at addressing various open-world video recognition tasks.
The reason why we choose these three video benchmarks is that,
these benchmarks are representative with quite different open domains in the downstream,
i.e.,
they are suitable to verify if the proposed PCA can boost such diversified and challenging open-world video recognition,
via multi-modal knowledge transfer of foundation models.

\noindent\textbf{Implementation Details}.
Common settings for the TinyVIRAT, ARID, and QV-PIPE datasets are listed below. 
We evaluated the performance of the PCA pipeline using five popular models: R(2+1)D\cite{r2p1d} for CNN-style models,
CLIP-ViT\cite{vit, clip} and Timesformer\cite{timesformer} for transformer-style models, as well as CLIP-3D and UniFormer\cite{uniformer} for CNN-transformer hybrid models. CLIP-3D is obtained by inserting 3D convolutions into ViT\cite{vit} encoder module following UniFormerV2\cite{uniformerv2}. It performs well in low-resolution recognition tasks as it can encode both CLIP pretraining knowledge and temporal information. 
Regarding model initialization, we used publicly available pretrained weights. The R(2+1)D model used weights pretrained on the IG65m\cite{ig65m} dataset. The CLIP-ViT and CLIP-3D models used CLIP\cite{clip} pretraining. The Timesformer\cite{timesformer} and UniFormer\cite{uniformer} models used Kinetics-400\cite{kinetics400} pretraining. 
As for the model hyperparameters,
we follow the original settings of different models\cite{clip, timesformer, uniformer, r2p1d}.
And we use AdamW\cite{adamw} optimizer with momentum $\beta_1, \beta_2 = 0.9, 0.999$, and cosine decay for  learning rate. 
All experiments were conducted on 8 NVIDIA RTX A6000 GPUs. 
Other different settings for each dataset are listed below.

In TinyVIRAT, 
the specific hyperparameters are as follows: the learning rate is set to 1e-5, the batch size is set to 72, the total number of training epochs is 90, the warmup epoch is set to 10, and the weight decay is set to 0.01. During training and testing, the images are resized to 224$\times$224 pixels using trilinear interpolation, and 8 frames are uniformly sampled. We use the F1-Score to evaluate the model performances. We use a super-resolution model (RealBasicVSR\cite{realbasicvsr}) to enhance video with crucial visual details.
Specifically,
we use the RealBasicVSR$\_\times$4 model from its official code. We uniformly sample 16 frames of each video and apply super-resolution to 4 times of the original resolution.

In ARID, most of our hyperparameters are consistent with Darklight\cite{darklight}. The learning rate is set to 1e-5, the batch size is 32, the warmup epochs is 5 and the total training epochs is 120. 
We uniformly sample 8 frames per video and apply the gamma transformation to these frames.
During training, we apply multi-scale crop and random horizontal flip augmentation. During testing, we use center-crop augmentation. The gamma correction\cite{dip} rate is set to 1.8, and the classification head has a dropout rate of 0.5. The Top-1 accuracy and Top-5 accuracy are used for evaluation.





In QV-Pipe, 
the learning rate, weight decay, batch size, warmup epochs, and total training epochs, are set to 1e-5, 0.2, 72, 5, and 120 respectively.
Since it has large domain shifts with fine-grained anomaly categories,
we used the Segment Anything (SAM) model\cite{segmentanything} to discover detailed semantic regions in such open-world scenes.
We uniformly sample 8 frames with the resolution of 224$\times$224. We use the segment anything model whose image encoder is MAE\cite{mae} pretrained ViT-H\cite{vit} as mentioned in the segment anything technical report\cite{segmentanything}. By average fusion of semantic segmentation masks into the original video frames, 
we highlight the critical regions in the pipe videos for visual knowledge extraction.
During training, multi-scale crop, color jitter, and random grayscale augmentation are applied. During testing, the center crop is used. The dropout rate for the classification head is 0.5. MAP (Mean Average Precision) is the evaluation metric.

\subsection{Comparison to the State of the Art}

\begin{table*}
\renewcommand{\arraystretch}{1.15}
\begin{minipage}[t]{0.5\textwidth}
\centering
\Large
\caption{Comparison with the state-of-the-art on TinyVIRAT\cite{tinyvirat}}
\label{tab:tinyvirat sota}
\resizebox{0.5\columnwidth}{!}{
\begin{tabular}{lc}
\toprule
\textbf{Model }                 & \textbf{F1-Score}         \\
\midrule
R(2+1)D\cite{r2p1d}               & 74.10\%          \\
CLIP-ViT\cite{clip}               & 74.70\%          \\
Timesformer\cite{timesformer}            & 77.40\%          \\
UniFormer\cite{uniformer}              & 77.80\%          \\

CLIP-3D\cite{clip}                & 78.80\%          \\
EVL\cite{frozenclip}                    & 79.33\%          \\
ST-Adapter\cite{stadapter}            & 79.60\%          \\ \midrule
PCA{\fontsize{7.0pt}{\baselineskip}\selectfont R(2+1)D}   & 77.10\%          \\

PCA{\fontsize{7.0pt}{\baselineskip}\selectfont Timesformer}     & 79.50\%          \\

PCA{\fontsize{7.0pt}{\baselineskip}\selectfont CLIP-ViT}       & 79.90\%          \\
PCA{\fontsize{7.0pt}{\baselineskip}\selectfont UniFormer}       & 80.40\%          \\
\textbf{PCA{\fontsize{7.0pt}{\baselineskip}\selectfont CLIP-3D}} & \textbf{82.28\%} \\
\bottomrule
\end{tabular}
    }
\end{minipage}\hspace{1em}
\begin{minipage}[t]{0.5\textwidth}
\caption{Comparison with the state-of-the-art on QV-Pipe\cite{videopipe}}
\label{tab:sota videopipe}
\centering
\large
\resizebox{0.5\columnwidth}{!}
{
\begin{tabular}{lc}

\toprule
\textbf{Model}                   & \textbf{mAP} \\
\midrule
R(2+1)D\cite{r2p1d}               & 25.67\%                          \\
UniFormer\cite{uniformer}               & 50.72\%                          \\
CLIP-3D\cite{clip}                 & 55.80\%                          \\
Timesformer\cite{timesformer}             & 62.84\%                          \\
EVL\cite{frozenclip}                     & 63.13\%                          \\
CLIP-ViT\cite{clip}                & 63.40\%                          \\
ST-Adapter\cite{stadapter}             & 63.42\%                          \\ \midrule
PCA{\fontsize{7.0pt}{\baselineskip}\selectfont R(2+1)D}         & 26.30\%                          \\
PCA{\fontsize{7.0pt}{\baselineskip}\selectfont UniFormer}         & 55.30\%                          \\
PCA{\fontsize{7.0pt}{\baselineskip}\selectfont CLIP-3D}            & 57.14\%                          \\
PCA{\fontsize{7.0pt}{\baselineskip}\selectfont Timesformer}       & 62.98\%                          \\

\textbf{PCA{\fontsize{7.0pt}{\baselineskip}\selectfont CLIP-ViT}  } & \textbf{65.42\%}                 \\
\bottomrule
\end{tabular}
}
 	\end{minipage}
\end{table*}
\noindent\textbf{TinyVIRAT. }We compared the current mainstream CNN\cite{r2p1d}, Transformer models\cite{clip, timesformer} and popular adaptation methods\cite{stadapter, frozenclip} on TinyVIRAT\cite{tinyvirat} as shown in Table \ref{tab:tinyvirat sota}. 
Among them, EVL\cite{frozenclip} and ST-Adapter\cite{stadapter} are representative video adapter models. We do not apply PCA on adapter methods because these models have been equipped with adapter modules.
It is observed that adaptation methods outperform other baselines. 
After incorporating the PCA method, improved results were achieved on R(2+1)D, UniFormer, Timesformer, CLIP-ViT, and CLIP-3D models compared to their respective baselines. 
Notably, the CLIP-ViT method, after PCA is incorporated, outperforms other adapter methods based on the CLIP model. The PCA{\fontsize{8pt}{\baselineskip}\selectfont CLIP-3D} model achieves the best results of 82.28\%.

\noindent\textbf{QV-Pipe. }The results of QV-Pipe dataset are reported in Table \ref{tab:sota videopipe}.
We can see that our proposed PCA{\fontsize{8pt}{\baselineskip}\selectfont CLIP-ViT} methods achieve SOTA performance compared to previous methods.
It outperforms the ST-Adapter method by +2.0\%.
Additionally,
we also showcase the results of other baseline methods\cite{uniformer, clip, timesformer, r2p1d} with PCA pipeline.
QV-Pipe is a complex multilabel dataset requiring global information for perception\cite{videopipe}. 
The pure convolution network can not generalize well, because convolution has limited capabilities to integrate global information\cite{uniformer}. 
Therefore, the performance of R(2+1)D\cite{r2p1d} is not as good as transformer-based models with global attention to QV-Pipe. By incorporating the PCA method, our model outperformed the respective baseline models, achieving state-of-the-art performance.

\begin{table}[!]
\caption{Comparison with the state-of-the-art on ARID\cite{arid}}
\label{tab:sota arid}
\centering
\resizebox{0.6\columnwidth}{!}
{
\begin{tabular}{lcc}
\Xhline{0.8pt}
\textbf{Model}                  & \textbf{Top-1 Acc}  & \textbf{Top-5 Acc} \\
\Xhline{0.6pt}
ST-Adapter\cite{stadapter}            & 72.67\%   & 97.51\%   \\
CLIP-ViT\cite{clip}               & 77.96\%   & 97.08\%   \\
Timesformer\cite{timesformer}            & 78.53\%   & 96.02\%   \\
CLIP-3D\cite{clip}               & 78.79\%   & 98.44\%   \\
EVL\cite{frozenclip}                    & 79.01\%   & 98.44\%   \\
UniFormer\cite{uniformer}              & 86.60\%   & 96.46\%   \\
DarkLight-ResNeXt\cite{darklight}      & 87.27\%   & 99.47\%   \\
R(2+1)D\cite{darklight}                & 90.45\%   & 98.11\%   \\
R(2+1)D-Light\cite{darklight}          & 91.55\%   & 99.51\%   \\
R(2+1)D-DarkLight-SA\cite{darklight}   & 94.04\%   & 99.87\%   \\
\hline
PCA{\fontsize{7.0pt}{\baselineskip}\selectfont CLIP-ViT}          & 81.89\%   & 97.75\%   \\
PCA{\fontsize{7.0pt}{\baselineskip}\selectfont Timesformer}       & 82.77\%   & 95.23\%   \\

PCA{\fontsize{7.0pt}{\baselineskip}\selectfont CLIP-3D}          & 83.89\%   & 97.83\%   \\
PCA{\fontsize{7.0pt}{\baselineskip}\selectfont UniFormer}        & 88.30\%   & 98.28\%   \\
\textbf{PCA{\fontsize{7.0pt}{\baselineskip}\selectfont R(2+1)D}} & \textbf{98.90\%}  & \textbf{99.96\%}      \\
\Xhline{0.8pt}
\end{tabular}
}
\end{table}

\noindent\textbf{ARID. }
We compared our approach with previous mainstream models in the ARID dataset in Table \ref{tab:sota arid}. 
After incorporating the PCA method, 
the performance of all models surpassed their corresponding baseline models. 
With the help of text and visual knowledge, 
PCA{\fontsize{8pt}{\baselineskip}\selectfont R(2+1)D} model achieves the best Top-1 Acc of 98.9\%.
Note that,
ARID has the limited number of training videos (less than 4k).
In general,
ViTs have more training parameters than CNNs,
e.g.,
ViT-Base has 86M model parameters,
while
R(2+1)D-18 has 33.3M model parameters.
Hence,
ViTs is more inclined to be overfitting than CNNs,
given such limited training data in ARID.
As a result,
R(2+1)D achieves a better performance than CLIP-ViT.
Additionally,
ARID refers to dark-illumination video recognition,
where action labels are fine-grained with high confusion.
This makes CLIP-ViT even worse on Top-1 Acc.
Hence,
the performance gap between R(2+1)D and CLIP-ViT in Top-1 Acc is larger than that in Top-5 Acc.

\subsection{Ablation Studies}
\begin{table*}[!htbp]
\caption{Multi-model Ablation for PCA pipeline.}
\label{tab:multimodal}
\centering
\resizebox{1\textwidth}{!}
{
    \begin{tabular}{l|cc|cccc}
        \Xhline{1.0pt}
        \multirow{3} * {\textbf{Model}}  
        ~ & \multicolumn{2}{c|}{\textbf{Addition Knowledge}} & \multicolumn{4}{c}{\textbf{Dateset}} \\ \cline{2-7}
        ~ & \multirow{2} * {\textbf{Visual Info}} & \multirow{2}*{\textbf{Text Info}} & \textbf{TinyVIRAT} & \multicolumn{2}{c}{ARID}  & \textbf{QV-Pipe}\\ \cline{4-7}
        ~ & ~ & ~ & \textbf{F1-Score} & \textbf{Top-1 Acc} & \textbf{Top-5 Acc} & \textbf{mAP}  \\ 
        \Xhline{0.8pt}
        \multirow{4} * {R(2+1)D\cite{r2p1d}} 
            & $\times$ & $\times$ & 74.10\% & 90.45\% & 98.11\% & 25.67\% \\
         ~ & $\checkmark$ & $\times$ & 76.00\% & 97.20\% & 98.90\% & 25.94\% \\
         ~ & $\times$ & $\checkmark$ & 76.20\% & 97.30\% & 99.00\% & 25.97\% \\
         ~ & \cellcolor{gray!20}{$\checkmark$} & \cellcolor{gray!20}{$\checkmark$} & \cellcolor{gray!20}{77.10\%} & \cellcolor{gray!20}{\textbf{98.70\%}} & \cellcolor{gray!20}{\textbf{99.40\%}} & \cellcolor{gray!20}{26.30\%} \\ \hline
         \multirow{4} * {UniFormer\cite{uniformer}}
         & $\times$ & $\times$ & 77.80\% & 86.60\% & 96.46\% & 55.30\% \\
         ~ & $\checkmark$ & $\times$ & 79.50\% & 87.86\% & 93.38\% & 56.12\% \\
         ~ & $\times$ & $\checkmark$ & 78.90\% & 87.09\% & 96.15\% & 56.51\% \\
         ~ & \cellcolor{gray!20}{$\checkmark$} & \cellcolor{gray!20}{$\checkmark$} & \cellcolor{gray!20}{80.40\%} & \cellcolor{gray!20}{88.30\%} & \cellcolor{gray!20}{98.28\%} & \cellcolor{gray!20}{56.86\%} \\ \hline
         \multirow{4} * {Timesformer\cite{timesformer}} 
         & $\times$ & $\times$ & 77.40\% & 78.53\% & 96.02\% & 60.55\% \\
         ~ & $\checkmark$ & $\times$ & 78.90\% & 82.25\% & 93.62\% & 61.19\% \\
         ~ & $\times$ & $\checkmark$ & 79.10\% & 81.57\% & 95.31\% & 62.27\% \\
         ~ & \cellcolor{gray!20}{$\checkmark$} & \cellcolor{gray!20}{$\checkmark$} & \cellcolor{gray!20}{79.50\%} & \cellcolor{gray!20}{82.77\%} & \cellcolor{gray!20}{95.23\%} & \cellcolor{gray!20}{62.98\%} \\ \hline
         \multirow{4} * {CLIP-ViT\cite{clip}} 
         & $\times$ & $\times$ & 74.70\% & 77.96\% & 97.08\% & 63.40\% \\
         ~ & $\checkmark$ & $\times$ & 77.80\% & 79.69\% & 95.63\% & 64.45\% \\
         ~ & $\times$ & $\checkmark$ & 78.20\% & 79.32\% & 96.15\% & 64.61\% \\
         ~ & \cellcolor{gray!20}{$\checkmark$} & \cellcolor{gray!20}{$\checkmark$} & \cellcolor{gray!20}{79.90\%} & \cellcolor{gray!20}{81.89\%} & \cellcolor{gray!20}{97.75\%} & \cellcolor{gray!20}{\textbf{65.17\%}} \\ \hline
         \multirow{4} * {CLIP-3D\cite{clip} }
         & $\times$ & $\times$ & 78.80\% & 78.79\% & 97.44\% & 55.80\% \\
         ~ & $\checkmark$ & $\times$ & 81.30\% & 82.05\% & 97.75\% & 56.57\% \\
         ~ & $\times$ & $\checkmark$ & 81.50\% & 83.01\% & 97.71\% & 56.79\% \\
         ~ & \cellcolor{gray!20}{$\checkmark$} & \cellcolor{gray!20}{$\checkmark$} & \cellcolor{gray!20}{\textbf{81.93\%}} & \cellcolor{gray!20}{83.89\%} & \cellcolor{gray!20}{97.83\%} & \cellcolor{gray!20}{57.14\%} \\
        \Xhline{1.0pt}
    \end{tabular}
}
\end{table*}

\noindent\textbf{Multimodal Ablation for PCA: } We insert one block of the adapt module to the baseline models for multimodal ablation. The ablation of the block number of the Adapt module is described in detail in the following subsection. According to Table \ref{tab:multimodal}, 
all models have gained performance improvement when incorporating only single-modal knowledge(either visual or text),
compared to their corresponding baselines.
This indicates that external knowledge, whether in the form of text or visual modalities, has a positive impact on improving the accuracy of the model.
Moreover, the integration of both visual and textual knowledge further boosts the models' performance, showing the effectiveness of our proposed method.

\begin{table}[!htbp]
\caption{Results of the different structures of the adapt module.}
\label{tab:structures}
\centering
\resizebox{0.6\columnwidth}{!}
{
\begin{tabular}{l|ccc}
\Xhline{1.0pt}
\multirow{2} * {\textbf{Structure}} & \textbf{TinyVIRAT} & \textbf{ARID} & \textbf{QV-Pipe}  \\ \cline{2-4}
~ & \textbf{F1-Score} & \textbf{Top-1 Acc} & \textbf{mAP}  \\
\Xhline{0.8pt}
Addition Module & 81.21\% & 97.80\% & 63.16\%  \\ 
Res-Cross Module & 80.80\% & 97.32\% & 63.80\%  \\ 
Res-Prompt Module & 81.32\% & 98.44\% & 64.29\%  \\ 
\rowcolor{gray!20}
Adapt Module & \textbf{81.93\%} & \textbf{98.70\%} & \textbf{65.17\%}  \\ 
\Xhline{1.0pt}
\end{tabular}
}
\end{table}

\noindent\textbf{Different Structures of Adapt Module: } 
To further verify the impact of different adapter module structures on the results.
We tested the above four different structures as shown in Fig.\ref{fig:structures} to validate
how they influence the final performance.
As shown in Table \ref{tab:structures}, the results of the addition module (Fig.\ref{fig:structures}(a)) and the Res-Cross module (Fig.\ref{fig:structures}(b)) are relatively low among the four structures.
Then, the Res-Prompt module (Fig.\ref{fig:structures}(c)) further boosts performance with the addition of a learnable query based on Res-Cross,
indicating that the learnable query can assist in integrating external information.
Finally, The Adapt module (Fig.\ref{fig:structures}(d)) achieves the best result on three datasets, showing the superiority over other structures.

\begin{table}[!htbp]
\caption{Results of different learnable query dimension.}
\label{tab:query dim}
\centering
\resizebox{0.57\columnwidth}{!}
{
    \begin{tabular}{c|ccc}
    \Xhline{1.0pt}
        \multirow{2}* {\textbf{Query Dim}} & \textbf{TinyVIRAT} & \textbf{ARID} & \textbf{QV-Pipe}  \\ \cline{2-4}
        ~ & \textbf{F1-Score} & \textbf{Top-1 Acc} & \textbf{mAP}  \\
        \Xhline{0.8pt}
        64 & 81.34\% & 97.54\% & 64.83\%  \\ 
        96 & 81.68\% & 97.65\% & 65.03\%  \\ 
        \rowcolor{gray!20}
        128 & \textbf{81.93\%} & \textbf{98.70\%} & \textbf{65.17\%} \\
        \Xhline{1.0pt}
    \end{tabular}
}
\end{table}

\noindent\textbf{The Dimension of Learnable Prompt Query: } 
To further investigate the relationship between the dimension of the learnable prompt query and the performance of the Adapt module,
we conducted the experiment
with various dimensions of the query.
As shown in Table \ref{tab:query dim},
with the increase of prompt query dimension, 
the performance of the model also improves slightly.
For the balance of computation cost and performance, we chose the 128 dimensions for our module.

\begin{table}[!]
\caption{Results of different insert block nums}
\label{tab:block num}
\centering
\resizebox{0.6\columnwidth}{!}
{
\begin{tabular}{c|ccc}
\Xhline{1.0pt}
    \multirow{2} * {\textbf{Block Num}} & \textbf{TinyVIRAT} & \textbf{ARID} & \textbf{QV-Pipe}  \\ \cline{2-4}
    ~ & \textbf{F1-Score} & \textbf{Top-1 Acc} & \textbf{mAP}  \\
    \Xhline{0.8pt}
    0 & 78.80\% & 90.45\% & 63.40\%  \\ 
    1 & 81.93\% & 98.70\% & 65.17\%  \\ 
    2 & 82.20\% & 98.71\% & 65.25\%  \\
    \rowcolor{gray!20} 
    3 & \textbf{82.28\%} & \textbf{98.90\%} & \textbf{65.42\%}  \\
    \Xhline{1.0pt}
\end{tabular}
}
\end{table}

\noindent\textbf{The Number of Inserted Adapt Blocks: }
To validate the ability of the adapt module to incorporate intermediate-level features into the backbone network, 
we extracted the intermediate layers of the visual feature extraction models and integrated them into the backbone. 
Specifically, the block num represents the total number of adapt modules inserted as shown in Fig.\ref{fig:adapt}. 
For the CLIP-3D, Timesformer, and CLIP-ViT models, we extracted the Residual Attention Blocks\cite{vit} features from the middle layers. 
For the UniFormer model, we extracted middle layer features from SABlock layers\cite{uniformer}. 
For the R(2+1)D model, we extracted intermediate features from BasicBlocks\cite{r2p1d}. 
We only introduced textual knowledge in the last block of the model. 
The results shown in Table \ref{tab:block num}, Block Num 3 achieves the best result on three datasets. With the increased number of inserted blocks, the results improve. 

\noindent\textbf{Results of Visual Enhancement: }The first row refers to training the baseline model using the source data. The second row represents training the baseline model using augmented visual enhanced data. The final row demonstrates the effect of incorporating multimodal knowledge into PCA. As shown in Table \ref{tab:vis}, introducing visual single-modal knowledge in PCA gets better performance, 
since the PCA visual incorporates information from both the source data and the visual enhanced data. 
Finally, incorporating knowledge allows the network to achieve superior results.

\begin{table}[t]
\caption{Results of visual enhancement}
\label{tab:vis}
\centering
\resizebox{0.75\columnwidth}{!}
{
\begin{tabular}{l | ccc}
    \Xhline{1.0pt}
    \multirow{2} * {\textbf{Model}} & \textbf{TinyVIRAT} & \textbf{ARID} & \textbf{QV-Pipe}  \\ \cline{2-4}
    ~ & \textbf{F1-Score} & \textbf{Top-1 Acc} & \textbf{mAP}  \\
    \Xhline{0.8pt}
    Baseline & 78.80\% & 90.45\% & 63.40\%  \\ 
    Baseline + Visual Enhance & 79.30\% & 95.46\% & 63.77\%  \\ 
    \rowcolor{gray!20} 
    \textbf{Our PCA} & \textbf{81.93\%} & \textbf{98.70\%} & \textbf{65.17\%}  \\
    \Xhline{1.0pt}
\end{tabular}
}
\end{table}
\begin{table}
\caption{Results of different thresholds in the Chat process}
\label{tab:threshold}
\centering
\resizebox{0.6\columnwidth}{!}
{
\begin{tabular}{c|ccc}
\Xhline{1.0pt}
    \multirow{2} * {\textbf{Threshold}} & \textbf{TinyVIRAT} & \textbf{ARID} & \textbf{QV-Pipe}  \\ \cline{2-4}
    ~ & \textbf{F1-Score} & \textbf{Top-1 Acc} & \textbf{mAP}  \\
    \Xhline{0.8pt}
    0.2 & 81.65 & 98.02  & 64.55\\ 
    0.5 & 82.28 & 98.90 & 65.42 \\ 
    0.7 & 81.21 & 98.25 & 64.06  \\
    \Xhline{1.0pt}
\end{tabular}
}
\end{table}

\noindent\textbf{Threshold in the Chat Stage:}
As its role refers to evaluate the prediction score (i.e., label probability),
we choose three thresholds from [0,1] for ablation in Table. \ref{tab:threshold}. 
We can see that,
the performance is the best with the threshold of 0.5.
With smaller threshold (e.g., 0.2),
the Chat stage tends to use the prompt of the predicted labels as external textual knowledge.
With smaller threshold (e.g., 0.7),
the Chat stage tends to use the caption of the original video as external textual knowledge.
Both cases would introduce unnecessary error,
due to the poor confidence of label predictions or the unsatisfactory caption of noisy videos.
Hence,
to balance prompt knowledge and caption knowledge,
the final threshold is chosen as 0.5 for the best result.


\noindent\textbf{Scalability of PCA:}
The goal of our PCA is to tackle open-world video recognition.
In such realistic scenarios,
the main challenge is how to tackle large domain shifts or unknown video scenes with limited/long-tailed data,
rather than how to model large-scale video data.
But still,
we perform our PCA on the large-scale Kinetics-400 \cite{kinetics400} to evaluate its scalability.
We here choose a strong backbone of video recognition, 
UnMasked Teacher (UMT) \cite{unmaskteacher}.
Specifically,
we use UMT-B/16 as our baseline and all the settings are the same as other datasets in the paper.
Under the pretraining setting of K710 (PT+FT) and the frame setting of $8\times3\times4$ \cite{unmaskteacher},
PCA$_{\text{UMT-B/16}}$ achieves a better TOP-1 accuracy than the baseline UMT-B/16 (87.8 vs. 87.4).
Note that,
the performance of Kinetics-400 is actually saturated.
Hence,
such an improvement of PCA$_{\text{UMT-B/16}}$ is reasonable to show its scalability.

\subsection{Visualization}
\begin{figure*}[t]
    \centering
     \includegraphics[width=\textwidth]{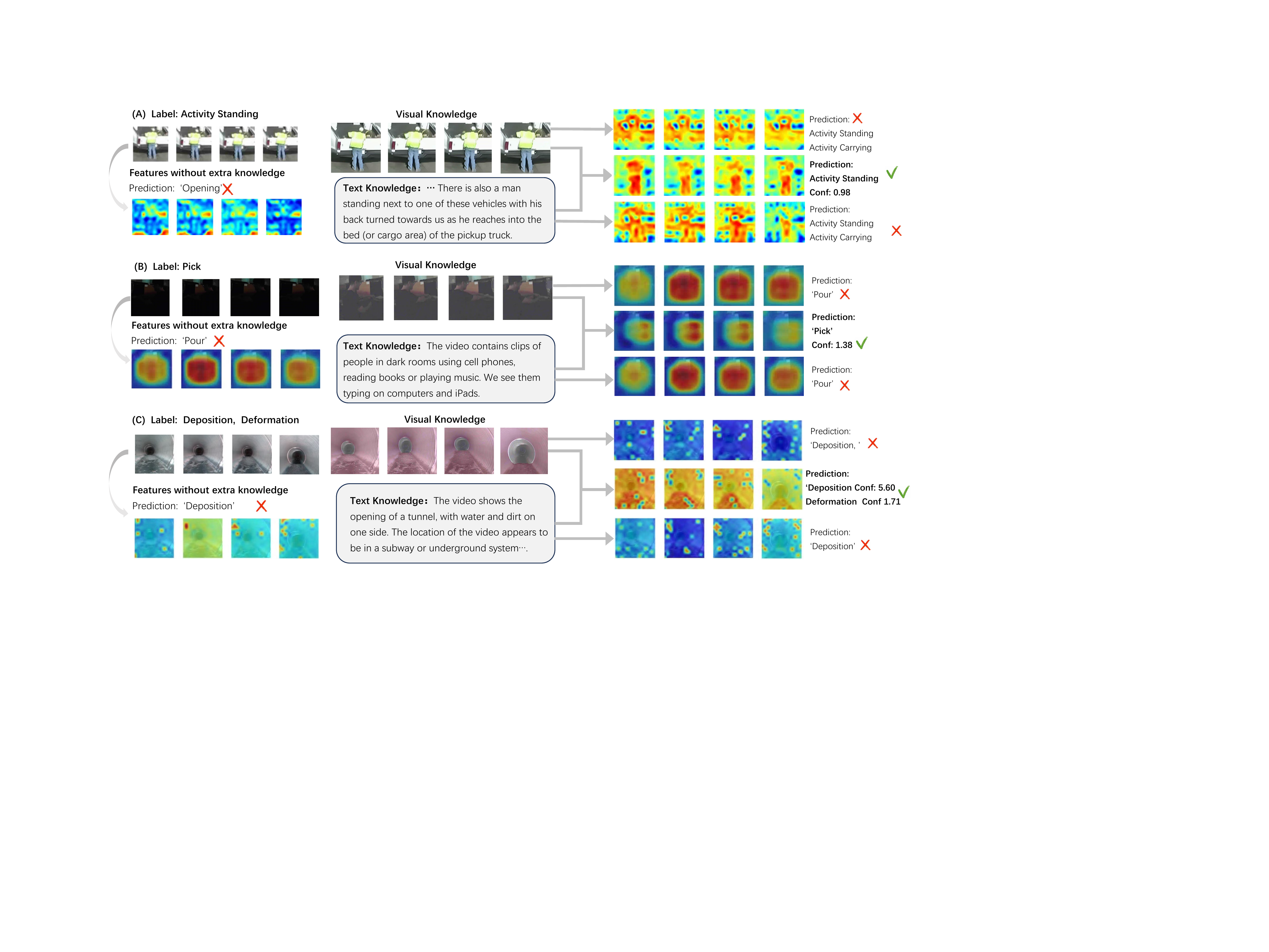}
    \caption{We use CAM\cite{cam} to visualize the PCA pipeline. For each case, the top left corner of each example shows four frames sampled from the original video, while the bottom left corner shows the features obtained without incorporating external knowledge. The images in the middle represents enhanced video frames. The text box contains the textual knowledge obtained by inputting the original video into the VideoChat foundation model. The upper right corner displays features that only include external visual knowledge, while the bottom right corner shows features that only include external textual knowledge. The middle section represents features obtained by incorporating both visual and textual knowledge.}
    \label{fig:vis}
\end{figure*}

We visualize the attention regions of the model using Grad-CAM\cite{cam} as shown in Fig.\ref{fig:vis}. After incorporating multi-model knowledge, the model gains the ability to correct the wrong answers provided by the baseline model.

(A) is from TinyVIRAT dataset. When there is no external knowledge, the model predicts the result as "Opening". 
After incorporating visual knowledge or textual knowledge alone, the model also can not predict the correct category. When the external knowledge includes both visual and textual modalities, the focus of the feature maps shifts to the body and feet and predict the correct category.

(B) is from the ARID dataset. baseline model makes incorrect judgments. 
The enhanced video restores hand movements in a dark environment.
Textual knowledge mentions the person using a phone, reading a book, or typing, indicating hand movements. 
With the incorporation of single-modal knowledge, the model still fails to determine the correct category. 
After incorporating multimodal knowledge, the model can infer comprehensively that the person is picking something and predict the right category.
Furthermore, the feature maps intuitively focus on the position of the hand.

(C) is from the QV-Pipe dataset. Without external knowledge and with only single-modal knowledge, the model cannot predict all class labels. 
Textual information indicates that this is a sewer with water and some dirt sediment. The model successfully obtains the correct result under the guidance of the semantic segmentation mask and the textual information.

\begin{figure*}[t]
   \centering
   \includegraphics[width=\textwidth]{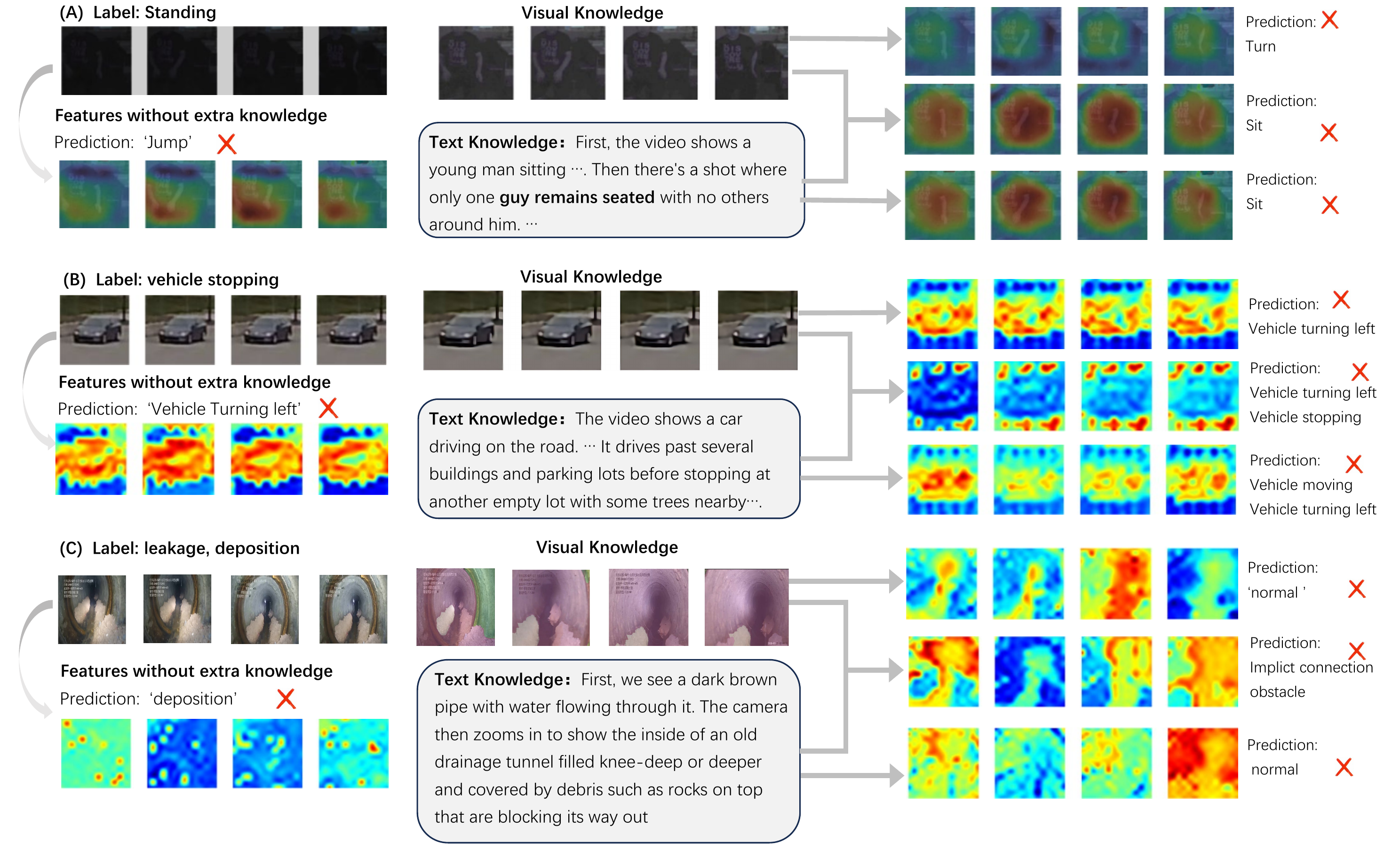}
   \caption{Visualization of the failure cases.}
\label{fig:wrong}
\end{figure*}

Additionally,
we visualize the failure cases of three datasets as shown in Figure.\ref{fig:wrong}.
The common problem in these cases is that,
the ground truth labels are highly confused with other labels in the datasets.
Hence,
visual and textual knowledge may be insufficient for these challenging cases.
It would be interesting to explore other knowledge (like motion, audio, depth) to further enhance our PCA in the future.

\section{Conclusion}
In this work, we propose a general knowledge transfer framework PCA, which utilizes a series of foundation models as knowledge containers and progressively incorporates and integrates external multimodal knowledge to enhance the performance of open-world video recognition. 
We enhance open-world videos to reduce domain gaps and extract external visual knowledge as Percept process. 
Then, in the Chat stage, we generate rich semantic information as external textual knowledge. 
Finally, we flexibly fuse external multimodal knowledge into the backbone for open-world recognition. 
Experimental results demonstrate the effectiveness of our PCA pipeline in open-world scenarios with low-resolution, bad illumination conditions, unusual video scenes, etc.
We have not explored how to combine this pipeline for more multi-modal tasks like detection, segmentation, grounding, etc. 
In future research, it could be explored how this method benefits more multi-modal tasks for open-world video recognition in scenarios like surveillance and robotics.


\bibliographystyle{elsarticle-num}
\bibliography{cas-refs} 
\end{document}